\documentclass{article}

\usepackage{balance} 
\usepackage{csquotes}
\usepackage{color}
\usepackage{todonotes}
\usepackage{glossaries}
\usepackage{cleveref}
\usepackage{url}

\newacronym{stbhv}{STBHV}{Schwartz's Theory of Basic Human Values}

\title{Value Engineering for Autonomous Agents}

\usepackage{authblk}

\author[1]{Nieves Montes}
\author[1]{Nardine Osman}
\author[1]{Carles Sierra}
\author[2]{Marija Slavkovik}

\affil[1]{Artificial Intelligence Research Institute (IIIA-CSIC)\\Barcelona, Spain}
\affil[2]{University of Bergen, Norway}
\affil[ ]{\texttt{\{nmontes,nardine,sierra\}@iiia.csic.es} \texttt{marija.slavkovik@uib.no}}

\begin{document}




\maketitle

\begin{abstract}
Machine Ethics (ME) is concerned with the design of Artificial Moral Agents (AMAs), i.e. autonomous agents capable of reasoning and behaving according to moral values. Previous approaches have treated values as labels associated with some actions or states of the world, rather than as integral components of agent reasoning. It is also common to disregard that a value-guided agent operates alongside other value-guided agents in an environment governed by norms, thus omitting the social dimension of AMAs. In this blue sky paper, we propose a new AMA paradigm grounded in moral and social psychology, where values are instilled into agents as context-dependent goals. These goals intricately connect values at individual levels to \emph{norms} at a collective level by evaluating the outcomes most incentivized by the norms in place. We argue that this type of normative reasoning, where agents are endowed with an understanding of norms' moral implications, leads to value-awareness in autonomous agents. Additionally, this capability paves the way for agents to align the norms enforced in their societies with respect to the human values instilled in them, by complementing the value-based reasoning on norms with agreement mechanisms to help agents collectively agree on the best set of norms that suit their human values. Overall, our agent model goes beyond the treatment of values as inert labels by connecting them to normative reasoning and to the social functionalities needed to integrate value-aware agents into our modern hybrid human-computer societies.
\end{abstract}



\section{Introduction}

Our society  is a sociotechnical system that includes human agents, humans that augment their abilities with computational  devices, and  artificial agents capable of increasing degrees of unsupervised action. Multi-agent systems as a discipline is concerned with identifying and implementing coordination-promoting interactions among agents. 

Agent coordination does not only require rationality and autonomy, it also requires an understanding of moral values --  ``the greater the freedom of a machine, the more it will need moral standards'' \cite{Picard1997}. Floridi and Sanders' \cite{Floridi2004} definition of moral agent clearly includes artificial agents: ``An action is said to be morally qualifiable if and only if it can cause moral good or evil. An agent is said to be a moral agent if and only if it is capable of morally qualifiable action''. How can we make artificial moral agents? 

The area that investigates how to build artificial moral agents that are capable of moral reasoning is machine ethics (ME). Machine ethics  ``is concerned with the behaviour of machines towards human users and other machines'' \cite{Anderson2007}. ME's goal is to enable automation and augmentation algorithms to uphold the ``right" values. Machine ethics is however, not directly concerned with how to integrate moral reasoning together with an agent's interactions. What values are and how are they sourced is also very much an open question. To attain a socio-technical society in which the collective and individual values are aligned, we need to be able to construct artificial moral agents that ``understand'' values. 






We here propose a new artificial moral agent paradigm that allows for human-selected values to be embedded in an artificial agent reasoning process. We follow moral psychology rather than moral philosophy when reaching a formal specification for ``value''. We motivate our reasoning and approach in Section~\ref{sec:background}, and present the approach in \Cref{sec:contribution}. There, we establish how agents can be endowed with the meaning of values and what view on norms we take. We also propose a mechanism for aligning values with norms that govern the behaviour in a shared environment. We conclude in \Cref{sec:conclusions} by discussing the novelty of our approach with respect to previous ones, and outlining directions for future work. 



\section{Background on values and norms}\label{sec:background}

\subsection{What are values}
In the broader domain of AI Ethics, the question of which values should be imposed on artificial agents (their use and development) dominates the landscape \cite{Hagendorff:2020,JobinIV2019}. However, the term ``value'' is often taken without being defined and not really linked to ethics. 

In moral philosophy, ethical values are typically defined in relation to ethical principles. Ethical principles are part of a normative theory that justifies or defends moral rules and/or moral decisions and reflects objective positions of right and wrong or what ``ought'' to be \cite{Fieser}. The IEEE 7000-2021 IEEE Standard Model Process for Addressing Ethical Concerns during System Design\footnote{\url{https://standards.ieee.org/ieee/7000/6781/}} defines ethical values as ``value in the context of human culture that supports a judgment on what is right or wrong''. Values, broadly understood as this, are difficult to operationalise for computational agents. As a result, researchers take many different approaches, which are hard to compare or evaluate. Moreover, such definition does not facilitate agent interaction either. Namely, the developed paradigms and prototypes of artificial moral agents are such that the agent does not include the moral values of others when reasoning about its own actions and states. 

In reinforcement learning, which is increasingly explored as an approach for attaining morally behaving agents \cite{ChaputDBGH21,Noothigattu2019,armstrong2015motivated}, we see a vast variety of formalisations (and quantifications) for values. In other machine ethics approaches, such as \cite{Liao2018,AndersonA14,DennisFSW16}, values are used as labels for actions, options or other elements of the reasoning process. 

To be able to build value ``aware'' agents, we need to formalise values as an element of the agent reasoning process. The choice of specialisation should allow qualification of the level of agent value ``awareness'' and agent interaction. 

\subsection{What are norms (and multi-agent systems)}\label{elinor}
Agents who interact in a shared environment will be subject to the norms regulating it. Norms are social mechanisms that steer human behaviour towards certain desired outcomes. Norms are responsible for structuring the interactions among agents and of their overall organisation. Historical examples show that communities of humans have been successful and sustainable when its members follow a set of self-crafted, well-defined norms \cite{Ostrom1990}. The reasons why a human follows a particular norm can come from a variety of sources, it can be because it is beneficial for them or perhaps because it is understood as a moral or professional obligation. 

The formal languages used to represent norms are very rich and varied: constraint-based \cite{DBLP:journals/aamas/Garcia-CaminoRSV09}, time-based \cite{DBLP:journals/sLogica/AgotnesHRSW09}, or event-based \cite{DBLP:books/ox/05/KowalskiS05}, just to name a few. In many cases, specially in human societies, they come in the form of natural language expressions. However, most representation languages, formal or not, are consequential in the sense that norms act via punishments or rewards. Certain human actions make humans incur in a cost, while others provide a benefit. 

Norms are a system-level construct, that apply to the society of agents as a whole. This feature of norms is a contrast with values, that reside on every individual agent. 





\section{Building Value-Aware Agents}\label{sec:contribution}
This section presents our proposal for an artificial value-aware agent. 
We call an agent value-aware when it has an explicit representation of the operational meaning of values, one that allows it to interpret the state of the MAS according to those values. Our main claim of this paper is as follows. If agents are value-aware, then they will be capable of reasoning about norms from the perspective of the value-alignment of those norms. In other words, they can analyse a set of norms in terms of the outcomes that it promotes (or the MAS behaviour it brings about) and the degree of alignment of those outcomes with the desired values. This opens the door to the creation and selection of norms from a moral perspective, and influencing the value-alignment of a MAS from within the system itself. 

This section is divided into three parts. The first presents our view on values, and how representing them through concrete goals opens the door for developing value aware agents. The second presents our view on norms. Finally, the third links values to norms, which provides the basis for reasoning about norms and MAS behaviour from the perspective of the values they are aligned with.

\subsection{What are values for value-aware agents}

As a starting point to build our artificial moral agent, we take the \acrfull{stbhv} \cite{Schwartz1992,Schwartz2012}. This is a well-established theory in moral and social psychology, which provides a definition for values, outlines the functions they serve in social life and hints at how value structures are organized. In addition, many of the features of values it outlines are compatible with other frameworks from the social sciences and humanities \cite{Rohan2000}.

\begin{figure}[b]
    \centering
    \includegraphics[width=0.8\linewidth]{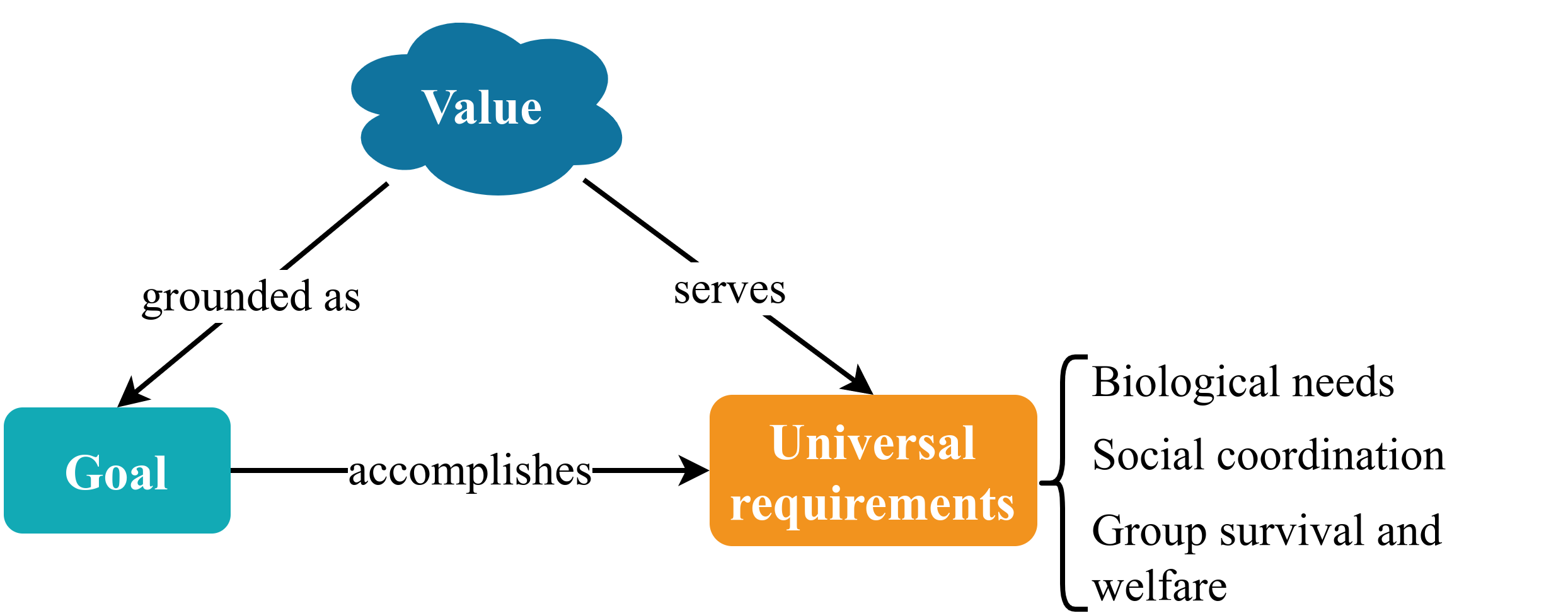}
    \caption{The three main components of \acrlong{stbhv}.}
    \label{fig:values-goal-function}
\end{figure}

Taking inspiration from the \acrshort{stbhv}, we start by establishing what values are and how they are made operational in humans. Schwartz acknowledges that values are \textquote{concepts or beliefs} that \textquote{transcend specific situations} \cite[p.~4]{Schwartz1992}. Despite their abstract nature, values are closely related to an individual's situation: \textquote{the primary content aspect of a value is the type of goal or motivational concern that it expresses. (...) values represent, in the form of conscious goals, three universal requirements of human existence to which all individuals and societies must be responsive (...)} \cite[p.~4]{Schwartz1992}. Like this, the \acrshort{stbhv} links values to two other concrete entities: the explicit goals that values motivate in specific situations and the ultimate functions that these goals seek to achieve. The relationships between values, goals and the requirements of human existence they fulfil appear in \Cref{fig:values-goal-function}.

We take these motivational goals from the \acrshort{stbhv} as the entities to makes values operational in a computational context. In other words, goals are the expression of the values that agents pursue, and they capture the meaning of a value in a particular situation or domain. Hence, the encoding of a goal into a software agent serves as a proxy for the value it is motivated by. While values are transcendental, their corresponding goals are \emph{context-dependent}. For example, in a professional context, the value ``gender equality'' should be grounded as equal recognition for equal work in terms of salary and career advancement. Meanwhile, in a domestic context, the same value is better reflected by the even division of domestic tasks.

Traditionally, the term ``goal'' in AI denotes a \emph{hard goal} with a clear-cut definition that is evaluated to either true or false \cite{Wooldridge1995}. We believe that this kind of dichotomous goals are not nuanced enough to reflect the complexities of values. Hence, we propose to encode value-motivated goals using fuzzy logic statements that can capture levels of satisfaction on a continuous scale \cite{Zadeh1975,Ramik2000}. Furthermore, this approach can also handle the relative importance of values (another feature identified in the \acrshort{stbhv}) by assigning \emph{satisfaction thresholds} to their corresponding goals (analogous to the graded desires defined in \cite{Casali2011}), so an agent does not seek to achieve a goal to its fullest, but to a \emph{satisfactory} extent. Just as value-motivated goals are context-dependent, so is the relative importance of values (i.e. the preference ordering over them), as a result, the thresholds that annotate their corresponding goals.

For example, if the value \emph{equality} is considered important in a context of wealth distribution, this value can be grounded as goal \emph{economic-equality} and its degree of truthfulness evaluated according to the Gini index computed from the agent's perception of wealth distribution. If the agent cares deeply about this value, its satisfaction threshold will be large (i.e. small Gini index), while if equality is low on its list of preferences its satisfaction threshold will be small (i.e. even a large Gini index will not spark the agent to act).

Now that we have an encoding of values as goals, we can discuss what we use these goals for. Here again, we take inspiration from the \acrshort{stbhv}, which essentially states that values operate as \emph{evaluation devices} through their goal proxies \cite{Schwartz1992}. These value-guided evaluations can be applied to a variety of constructs, such as actions, plans (i.e. sequences of actions), states, outcomes, or a combination of the above. Therefore, Schwartz's theory does not explicitly commit to a deontological or utilitarian position. In particular, value-motivated goals can also evaluate \emph{norms}, i.e. patterns or directives on behaviour, either through the actions they prescribe/forbid or through the outcomes that their implementation brings about. The relationship between norms and values is central to our proposal and is detailed in \Cref{subsec:claim1}.

One point that the \acrshort{stbhv} does not address in depth is the \emph{origin of values}. From the universal requirements for human existence that values serve (see \Cref{fig:values-goal-function}), one can infer that values are a consequence of evolution, and that they alleviate the cognitive load of having to continuously think in terms of sheer survival. In the context of artificial software agents, such concerns do not apply. Nonetheless, values, their motivating goals and their relative importance has to originate somewhere. For the time being, however, our proposal remains agnostic with respect to the value elicitation process. We are concerned with the inner operation of value-aware agents, and not, for the moment, with the specifics on \emph{how} and \emph{from whom} value-motivated goals are queried from.










\subsection{What are norms for value-aware agents}
We consider norms in this paper at this level of abstraction: norms, by acting on rewards and punishments, make certain environment transitions, resulting from agent actions, more probable than others. This view of ``institutional'' norms is rather general and includes ``simple'' norms such as those whose consequences are deontic (e.g. prohibition, permission, or obligation) operators over actions when a given pre-condition is satisfied.


\subsection {Connecting Values \& Norms}\label{subsec:claim1}

As illustrated above, norms govern behaviour: they incentivize behaviour to go in a particular direction. As such, we argue that norms have a central role as the primary value-promoting mechanisms. Norms can, if carefully designed, facilitate the fulfilment to a large extent of the goals that ground the meaning of values in the environment where the agents are operating. 

When implementing a new norm (or set of norms) leads to an outcome that is viewed as highly positive with respect to some value, we say that the norm {\em is aligned with respect to that value}. Hence, the relationship between norms and values is \emph{consequential} in nature. A norm is not moral in itself, it is so to the extent that the effects it brings about in the society agree with the members' values, represented in the form of goals.

\begin{figure}[t]
    \centering
    \includegraphics[width=0.6\linewidth]{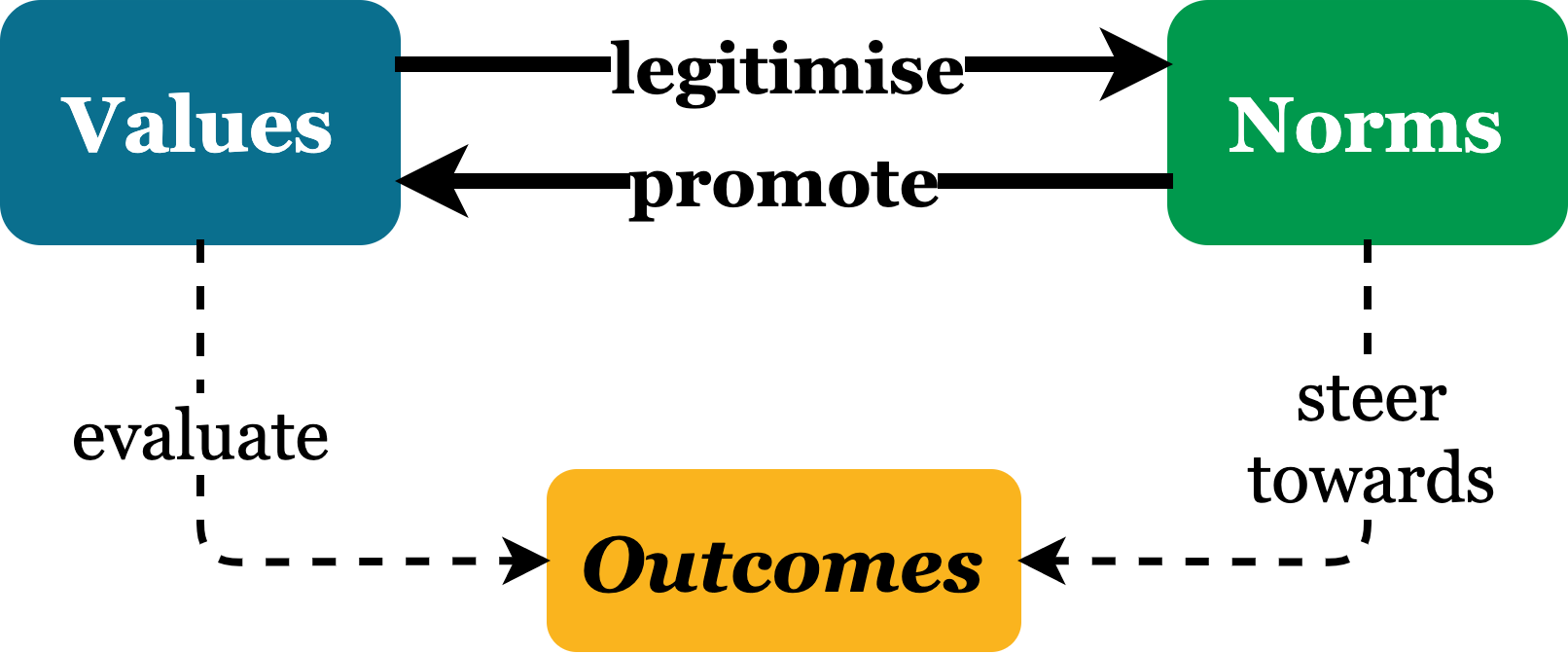}
    \caption{Relationship between values and norms through outcomes.}
    \label{fig:values-norms-relationship}
\end{figure}

Figure \ref{fig:values-norms-relationship} represents the relationship between the two entities in schematic form. At the surface, values and norms form a feedback loop: values legitimise the enforced norms, and norms promote values when enforced. At a more fundamental level, the two are linked by the outcomes that norms steer the system towards and that are favourably evaluated in regard to values. 

Ensuring AI is aligned with human values has been argued to be one of the main concerns for ethical AI~\cite{DignumBBCCDGHKL18}. To address this problem, and given the role of norms in driving behaviour, some works on automated norms synthesis have emerged, focusing on selecting norms on the basis of the moral values they support, see e.g. \cite{Serramia2018,Serramia2020}. Another line of work is Value-Sensitive Design (VSD), where the composition of morally adequate technical norms is hand-crafted by a human designer, which is made outside the multiagent system.



In our view, it should be the autonomous agents who attempt to align norms towards the values that the human designer has instilled in them. To the best of our knowledge, only \cite{MorrisMartin2019} has proposed an architecture for achieving the endogenous emergence of prescriptive norms through the participation of the agents. However, we are not aware of any follow-up on that work.

We propose that prescriptive norms,  with their explicit representation and syntax, should be handled by the agents populating the system, and evaluated by leveraging their understanding of values (i.e. the goals that values are grounded into) as evaluating devices. The designer, tasked with programming the agents, is not in charge of coding the technical norms directly. He/she is responsible, however, for including the necessary mechanisms that agents can resort to when crafting norms, figuring out the most probable outcomes that they lead to and ethically evaluating them.

The agents populating the system are instilled with values by a human who grounds their meaning as persistent goals, so that humans have complete control over the meaning of values. Every agent can be provided with its own, potentially different, version of grounding goals for the same value, so different agents might have conflicting value-grounding goals. Norms, on the other hand, are designed for the collective, with the objective of mediating the behaviour of the agent society as a whole. As illustrated above, our stance is that agents are responsible for aligning norms with the values that the human designer has instilled in them. 


Agreement technologies could be applied here, making use of computational social choice, argumentation, and negotiation mechanisms \cite{Baum20}. Of course, this may result in 
collective norms conflicting with some individual values, but representative of the values in the system overall.
The degree of value alignment of a norm can be assessed by the agent either analytically, or via simulations. The analytical approach involves formal reasoning to analyse the outcome of the behaviour induced by the norm. The alternative approach is to run simulations that would allow the agent to observe the outcome of norms. In both cases, the objective is to evaluate to what extent the value-grounding goals are satisfied in that outcome, which represents the degree of value-alignment of the norm. 

Our proposal, in summary, discusses how agents can collectively select the norms governing their interactions in such a way that ensures value-aligned behaviour. This is achieved through the agent's capability to reason about the value-alignment of norms, and hence, morally relating norms or sets of norms with values. This can be understood as empowering agents by making them value-aware. The capability of analysing norms from a moral perspective results in making value-aware decisions when creating, selecting, combining norms, or even deciding to abide by or break norms. This is the essential claim of this paper.



\section{Conclusion}\label{sec:conclusions}
Integrating ethical values in artificial agents becomes a necessity with the increase of the autonomy of the artificial agents. Current approaches to constructing artificial moral agents are based on implementing a specific moral theory or (typically reinforcement) learning moral behaviour. Ethical values are used as a syntactical construct, not as a part of the agent reasoning process. Specifically, values are used as  labels that human programmers assign to particular options or plans. Alternatively, values have also  been used to set human-given constraints, or regimented norms, on the available artificial agent actions, limiting the behaviour of the agent, as in \cite{DennisFSW16,Arkin2012}.  In addition, artificial moral agent ``design'' typically also omits the social aspects of agency \cite{Nallur20,Tolmeijer2020}, namely that agents exist in environments that are governed by laws and norms, which they share with others who are also moral agents.  

We propose a change in the agent paradigm that allows us to specify values as semantic constructs that can not only be used in the agent reasoning process, but also used to adjust the norms of the shared environment. 
In addition to reasoning about norms, we also aspire to agents that being value-aware, can reason about their own actions, and when to follow or break norms. 

What we have not discussed in the scope of this paper is the embedding of agents with value-enriched theory of mind. A value-enriched theory of mind would allow an agent to directly reason about other agents' behaviour, which would in turn lead to more value-aware interactions. In our proposed paradigm, the agent reasons only about norms, which are influenced by the values of the other agents. 
However, we leave the discussion of reasoning about agent actions to another paper.





\bibliographystyle{acm}
\bibliography{references}


\end{document}